\documentclass[runningheads]{llncs}

\usepackage{eccv}

\usepackage{eccvabbrv}

\usepackage{graphicx}
\usepackage{booktabs}

\usepackage[accsupp]{axessibility}  %
\usepackage{geometry}
\usepackage{hyperref}

\usepackage{orcidlink}
\usepackage{multirow}

\newif\ifreview

\ifreview
\newcommand{\model}{VEdit\xspace}
\newcommand{\fullmodel}{VEdit\xspace}
\else
\newcommand{\model}{EVE\xspace}
\newcommand{\fullmodel}{Emu Video Edit (EVE)\xspace}
\fi

\newcommand{\fullmethod}{Factorized Diffusion Distillation\xspace}
\newcommand{\method}{FDD\xspace}

\newif\ifcomments
\ifcomments
    \providecommand\yuval[1]{[\textcolor{blue}{Yuval: {#1}}]}
    \providecommand\shelly[1]{[\textcolor{purple}{Shelly: {#1}}]}
    \providecommand\adam[1]{[\textcolor{magenta}{Adam: {#1}}]}
    \providecommand\amit[1]{[\textcolor{violet}{Amit: {#1}}]}
    \providecommand\uriel[1]{[\textcolor{green}{Uriel: {#1}}]}
    \providecommand\yaniv[1]{[\textcolor{red}{Yaniv: {#1}}]}
    \providecommand\ap[1]{[\textcolor{orange}{Adam: {#1}}]}
    \providecommand\todo[1]{[\textcolor{purple}{TODO: {#1}}]}
\else
    \providecommand{\yuval}[1]{}
    \providecommand{\shelly}[1]{}
    \providecommand{\adam}[1]{}
    \providecommand{\amit}[1]{}
    \providecommand{\uriel}[1]{}
    \providecommand\ap[1]{}
    \providecommand{\todo}[1]{}
    \providecommand{\yaniv}[1]{}\providecommand{\reviewer}[1]{}
\fi

\begin{document}

\title{Video Editing via \fullmethod} 
\titlerunning{Video Editing via \fullmethod}

\author{Uriel Singer\inst{*} \and
Amit Zohar\inst{*} \and
Yuval Kirstain \and
Shelly Sheynin \and
Adam Polyak \and
Devi Parikh \and
Yaniv Taigman
}

\authorrunning{Singer et al.}

\institute{Meta AI}

\maketitle

\begin{abstract}

We introduce \fullmodel, a model that establishes a new state-of-the art in video editing without relying on any supervised video editing data. 
To develop \model we separately train an image editing adapter and a video generation adapter, and attach both to the same text-to-image model. 
Then, to align the adapters towards video editing we introduce a new unsupervised distillation procedure, \fullmethod.
This procedure distills knowledge from one or more teachers simultaneously, without any supervised data. 
We utilize this procedure to teach \model to edit videos by jointly distilling knowledge to (i) precisely edit each individual frame from the image editing adapter, and (ii) ensure temporal consistency among the edited frames using the video generation adapter.
Finally, to demonstrate the potential of our approach in unlocking other capabilities, we align additional combinations of adapters.

\end{abstract}

\ifreview
\else
\renewcommand{\thefootnote}{\fnsymbol{footnote}}  %
\footnotetext[1]{Equal contribution.}
\renewcommand{\thefootnote}{\arabic{footnote}}  
\fi
\section{Introduction}
\label{sec:intro}

The increasing usage of video as online content has led to a rising interest in developing text-based video editing capabilities.
However, due to the scarcity of supervised video editing data, developing such capabilities has proven to be challenging.
To address this challenge, the research community has mostly focused on training-free methods~\cite{Geyer2023TokenFlowCD,vid2vid-zero,yang2023rerender,text2video-zero,2312.04524}.
Unfortunately, these methods thus far appear to be limited both in terms of performance, and in the range of editing capabilities that they offer (Fig.~\ref{fig:qualitative_tgve}).

\begin{figure*}[htp!]
\centering
\includegraphics[width=.95\linewidth]{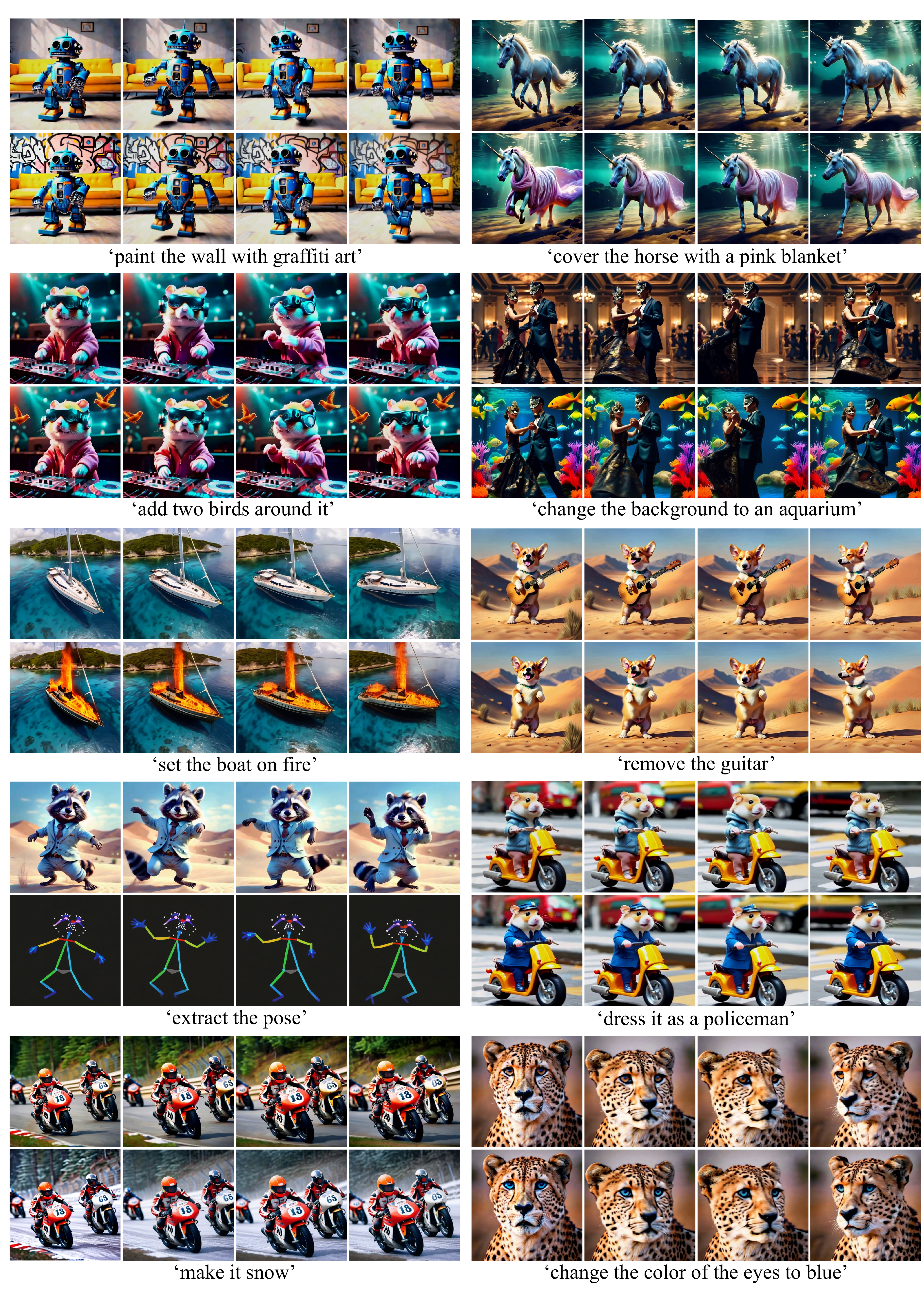}
\caption{\model is a text-guided video editing model that enables various editing tasks.}
\label{fig:method_sample}
\end{figure*}

Therefore, we introduce a new approach that allows us to train a \textit{state-of-the-art} video editing model \textit{without} any supervised video editing data. 
The main insight behind our approach is that we can decouple our expectations from a video editing model into two distinct criteria: (i) precisely edit each individual frame, and (ii) ensure temporal consistency among the edited frames.

\begin{figure}[t]
\centering
\includegraphics[width=0.9\linewidth]{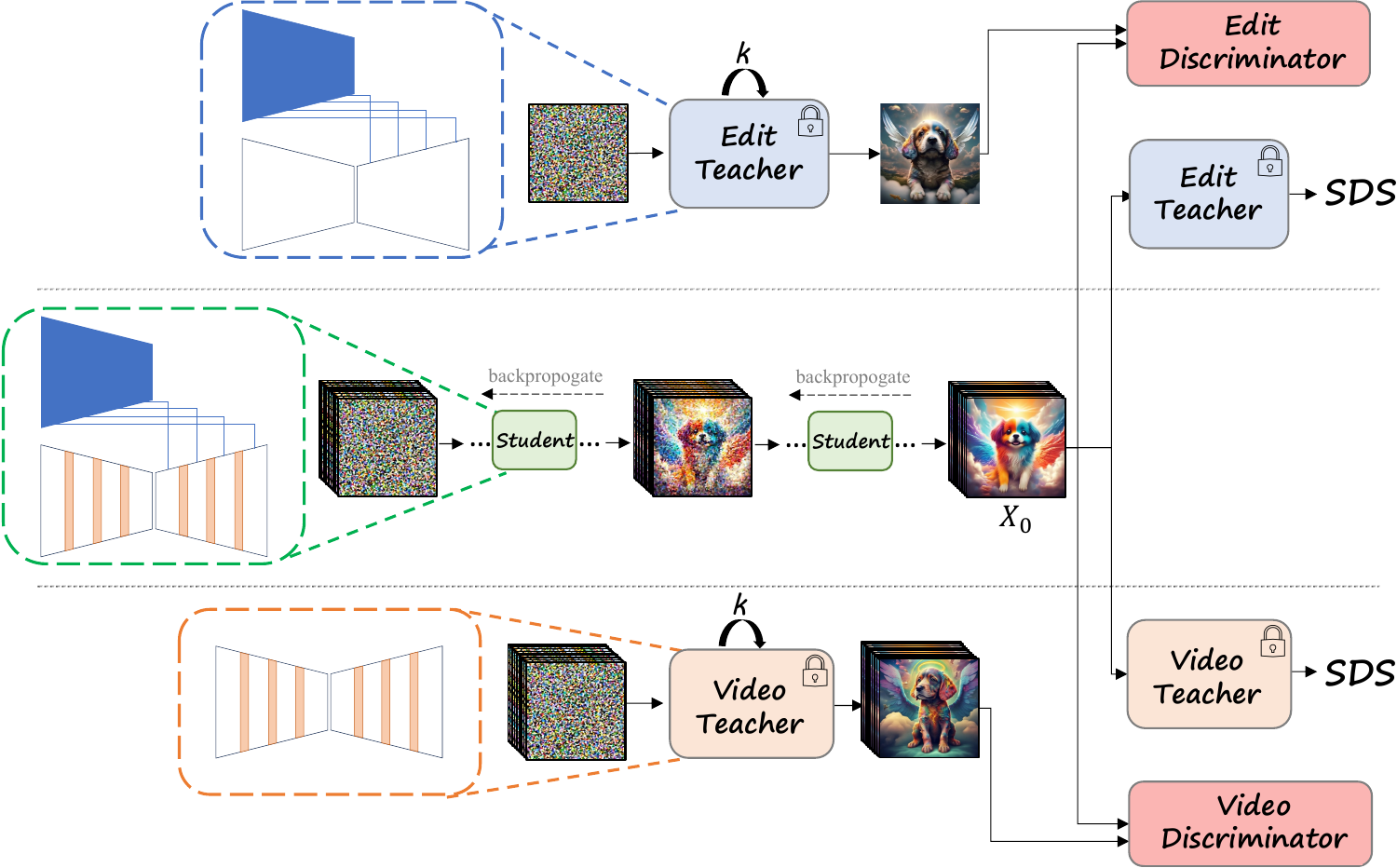}
\caption{Model architecture and alignment procedure. 
We train an adapter for image editing~(in blue) and video generation~(in orange) on top of a shared text-to-image backbone. Then, we create a student network by stacking both adapters together on the shared backbone~(in green) and align the two adapters. 
The student is trained using (i) score distillation from each frozen teacher adapter~(marked as SDS), (ii) adversarial loss for each teacher~(in pink). SDS is calculated on samples generated by the student from noise and the discriminators attempt to differentiate between samples generated by the teachers and the student.} 
\label{fig:arch}
\vspace{-0.3cm}
\end{figure}

Leveraging this insight we follow two phases. 
In the first phase, we train two separate adapters on top of the same frozen text-to-image model; an image editing adapter, and a video generation adapter. 
Then, by applying both adapters simultaneously we enable limited video editing capabilities.
In the second phase, we introduce a new unsupervised alignment method, \fullmethod~(\method), that drastically improves the video editing capabilities of our model. 
\method assumes a student model and one or more teacher models.
We employ the adapters as teachers, and for our student, we utilize trainable low-rank adaptation (LoRA) weights on top of the frozen text-to-image model and adapters. 
At each training iteration \method generates an edited video using the student.
Then, it uses the generated video to provide supervision from all teachers via Score Distillation Sampling~\cite{dreamfusion} and adversarial losses~\cite{Goodfellow2022GenerativeAN} 
(Fig.~\ref{fig:arch}).

The resulting model, \fullmodel, sets state-of-the-art results on the Text Guided Video Editing~(TGVE) benchmark~\cite{wu2023cvpr}.
Additionally, we improve two aspects of the evaluation protocol that was set in TGVE.
First, we utilize the recent ViCLIP~\cite{wang2023internvid} model to introduce additional automatic metrics that are temporally aware.
Second, we expand the TGVE benchmark to facilitate important editing tasks like adding, removing, and changing the texture of objects in the video. 
Importantly, \model also exhibits state-of-the-art results when tasked with these additional editing operations (Tab.~\ref{tab:eval_video_editing_tgve}). 

Notably, our approach can theoretically be applied to any arbitrary group of diffusion-based adapters. 
To verify that this holds in practice, we utilize our approach to develop personalized image editing models by aligning an image editing adapter with  different LoRA adapters~(Fig.~\ref{fig:additional_combination}). 

In summary, our method utilizes an image editing adapter and a video generation adapter, and aligns them to accommodate video editing using an unsupervised alignment procedure. 
The resulting model, \model, exhibits state-of-the-art results in video editing while offering diverse video editing capabilities. 
Furthermore, we extend the evaluation protocol for video editing by suggesting additional automatic metrics, and augment the TGVE benchmark with additional important editing tasks.
Finally, we verify that our approach can be used to align other adapters, and therefore, holds the potential to unlock new capabilities. %

\vspace{-0.1cm}
\section{Related Work}
\label{sec:related_work}
\vspace{-0.1cm}
The lack of supervised video editing data poses a major challenge in training precise and diverse video editing models.
A common strategy to address this challenge is via training-free solutions.
Initial work proposed the use of Stochastic Differential Editing (SDEdit)~\cite{sdedit}.
This approach performs image editing by adding noise to the input image and then denoising it while conditioning the model on a caption that describes the edited image.
Recently, several video foundation models, such as Lumiere~\cite{BarTal2024LumiereAS} and SORA~\cite{videoworldsimulators2024}, showcased examples in which they utilize SDEdit for video editing.
While this approach can preserve the general structure of the input video, adding noise to the input video results in the loss of crucial information, such as subject identity and textures.
Hence, SDEdit may work well when attempting to change the style of an image, but by design, it is unsuitable for \textit{precise} editing.

A more dominant approach is to inject information about the input or generated video from key frames via cross-attention interactions~\cite{Ceylan2023Pix2VideoVE,Geyer2023TokenFlowCD,Wu2023FairyFP,vid2vid-zero,yang2023rerender,text2video-zero,2312.04524,li2023vidtome,ma2023maskint,yatim2023space}.
Another common strategy is to extract features that should persist in the edited video, like depth maps or optical flow, and train the model to denoise the original video while using them~\cite{Liang2023FlowVidTI,Yan2023MotionConditionedIA,Esser2023StructureAC}. 
Then, during inference time, one can predict an edited video while using the extracted features to ensure faithfulness to the structure or motion of the input video.
The main weakness of this strategy is that the extracted features may lack information that should persist (e.g. pixels of a region that should remain intact), or hold information that should be altered (e.g. if the editing operation requires adding new motion to the video).
Consequently, the edited videos may still suffer from unfaithfulness to the input video or editing operation.

To improve faithfulness to the input video at the cost of latency, some works~\cite{wu2023tune,yang2023nvedit} invert the input video using the input caption.
Then, they generate a new video while using the inverted noise and a caption that described the output video. 
Another work~\cite{Cheng2023ConsistentVT} adapts the general strategy of InstructPix2Pix~\cite{brooks2022instructpix2pix} to video editing, which allows them to generate and train a video editing model using synthetic data.
While this approach seems to be effective, recent work in image editing~\cite{emu-edit} shows that Prompt-to-Prompt~\cite{hertz2022prompt} can yield sub-optimal results for various editing operations. %

In this paper we deviate from prior work. 
Instead, we distill distinct video editing capabilities from an image editing teacher and a video generation teacher.
Similarly to the Adversarial Diffusion Distillation (ADD)~\cite{Sauer2023AdversarialDD} loss, our approach involves combining a Score Distillation Sampling~\cite{dreamfusion} loss and an adversarial loss~\cite{Goodfellow2022GenerativeAN}.
However, it significantly differs from ADD. 
First, our method is unsupervised, and thus generates all data that is used for supervision rather than utilizing a supervised dataset. 
Second, we use distillation to learn a new capability, rather than reduce the number of required diffusion steps. 
Third, we learn this new capability by factorizing the distillation process and leverage more than one teacher model in the process.

\section{Method}
\label{sec:method}
The key insight behind our approach is that video editing requires two main capabilities: (1) precisely editing images, and (2) ensuring temporal consistency among generated frames.
In Sec.~\ref{videoadapter} we detail how we develop a dedicated adapter for each capability.
Next, we describe how our final architecture combines the adapters to enable video editing.
Finally, in Sec.~\ref{alignment} we introduce \fullmethod (\method), our method to align the adapters.
In Fig.~\ref{fig:arch} we provide an illustration of our model's architecture and \method.

\vspace{-0.1cm}
\subsection{Architecture}\label{subsec:arch}
\vspace{-0.1cm}
Our architecture involves stacking an image editing adapter and a video generation adapter on top of the same text-to-image backbone.
We employ the latent diffusion model, Emu~\cite{emu}, as our backbone model, and denote its weights with $\theta$.
We next describe how we develop and combine the different components to enable video editing.

\vspace{-0.2cm}
\subsubsection{Video Generation Adapter}
\label{videoadapter}
For the video generation adapter we make use of Emu Video~\cite{emu-video}, a text-to-video~(T2V) model that consists of trained temporal layers on top of a frozen Emu model. We thus consider the temporal layers as the video adapter.
Formally, text-to-video model output is denoted as $\hat{x}_{\rho}(x_s, s, c_{out})$, where $\rho=[\theta, \theta_{video}]$ are the text-to-image and video adapter weights, $x_s$ is a noisy video sample, $s$ is the timestep, and $c_{out}$ is the output video caption.
\vspace{-0.4cm}
\subsubsection{Image Editing Adapter}
\label{imageadapter}
To create an image editing adapter, we train a ControlNet~\cite{controlnet} adapter, with parameters $\theta_{edit}$, on the training dataset developed to train Emu Edit~\cite{emu-edit}.
We follow the standard practice of ControlNet training, and initialize the adapter with copies of the down and middle blocks of the text-to-image model.
During training we condition the text-to-image model on the output image caption, while using the input image and the edit instruction as inputs to our ControlNet image editing adapter.
Hence, we can denote the output of the image editing model with $\hat{x}_{\psi}(x_s, s, c_{out}, c_{instruct}, c_{img})$, where $\psi=[\theta, \theta_{edit}]$ are the text-to-image and image editing adapter weights, $x_s$ is a noisy image sample, $s$ is the timestep, $c_{out}$ is the output image caption, $c_{instruct}$ is the textual edit instruction, and $c_{img}$ is the input image we wish to edit.

\vspace{-0.2cm}
\subsubsection{Combining The Adapters}
\label{zeroshot}
To enable video editing capabilities we attach both of the adapters simultaneously to the text-to-image backbone~\footnote{To enable the editing adapter to process videos, we stack the frames independently as a batch.}.
Formally, we aim to denoise a noisy edited video $x_s$, using an input video $c_{vid}$, editing instruction $c_{instruct}$, and an output video caption $c_{out}$. 

Notably, when attaching the image editing adapter alone, the resulting function will process each frame independently.
Therefore, each frame in the predicted video should be precise and faithful to the input frame and editing instruction, but lack consistency with respect to the rest of the edited frames.
Similarly, when attaching the video generation adapter alone, the resulting function will generate a temporally consistent video that is faithful to the output caption, but not necessarily faithful to the input video.

When combining both adapters with the shared text-to-image backbone, the resulting function is $\hat{x}_\eta(x_s, s, c_{out}, c_{instruct}, c_{vid})$, where $\eta=[\theta, \theta_{edit}, \theta_{video}]$.
This formulation should enable editing a video that is both temporally consistent and faithful with respect to the input.
In practice, we observe that even though this ``plug-and-play'' approach enables video editing capabilities, it still includes significant artifacts, as we show in Sec.~\ref{sec:ablations}.
\vspace{-0.2cm}
\subsubsection{Final Architecture}
\label{finalarch}
As the necessary knowledge already exists in the adapters, we expect a small alignment to be sufficient. Hence, we keep the adapters frozen and utilize low-rank adaptation (LoRA)~\cite{lora} weights $\theta_{align}$ over the text-to-image backbone. Our final architecture becomes $\phi=[\theta, \theta_{edit}, \theta_{video}, \theta_{align}]$.
We describe in the following section how we train $\theta_{align}$ to improve the video editing quality of our model.

\subsection{\fullmethod}
\vspace{-0.1cm}
\label{alignment}
To train $\theta_{align}$ and align the adapters without supervised video editing data, we propose a new unsupervised distillation procedure, \fullmethod (\method).
In this procedure we freeze both adapters, and jointly distill their knowledge into a video editing student. 
Since our approach cannot assume supervised data, we only collect a dataset for the inputs.
Each data point in our dataset consists of $y = (c_{out}, c_{instruct}, c_{vid})$, where $c_{out}$ is an output video caption, $c_{instruct}$ is the editing instruction, and $c_{vid}$ is the input video.
We provide further details on this dataset in Sec.~\ref{sec:experiments}.

In each iteration of \method, we first utilize the student to generate an edited video $x'_0$ using a data point $y$ for $k$ diffusion steps (see Sec.~\ref{impl_details} for more details). 
Importantly, we will later backpropagate the loss through all of these diffusion steps.
We then apply Score Distillation Sampling (SDS)~\cite{dreamfusion} loss using each teacher. 
Concretely, we sample noise $\epsilon$ and a time step $t$, and use them to noise $x'_0$ into $x'_t$.
Then, we task each of the teachers to predict the noise from $x'_t$ independently. 
For a teacher $\hat{\epsilon}$, the SDS loss is the difference between $\epsilon$ and the teacher's prediction:
\begin{gather*}
    \mathcal{L}_\text{SDS}(\hat{x}) = \mathbb{E}_{x_0',t,\epsilon}[c(t)sg(\hat{\epsilon}(x'_t,t)-\epsilon)x_0'], 
\end{gather*}
where $c(t)$ is a weighting function, 
and $sg$ indicates that the teachers are kept frozen. The metric is averaged over student generations $x_0'$, sampled timesteps $t$ and noise $\epsilon$.
Plugging in the edit and video teachers, the loss becomes
\begin{gather*}
    \mathcal{L}_\text{SDS-Edit} = \mathcal{L}_\text{SDS}(\hat{x}_\psi), 
    \mathcal{L}_\text{SDS-Video} = \mathcal{L}_\text{SDS}(\hat{x}_\rho),
    \label{eq:vggface}
\end{gather*}
For brevity, we omit input conditions from $\hat{x}_\phi,\hat{x}_\psi,\hat{x}_\rho$.  
Each teacher provides feedback for a different criterion: the image editing adapter for editing faithfully and precisely, and the video generation adapter for temporal consistency. 

Similar to previous works employing distillation methods~\cite{meng2023distillation}, we observe blurry results, and therefore utilize an additional adversarial objective~\cite{Goodfellow2022GenerativeAN} for each of the teachers, akin to Adversarial Diffusion Distillation (ADD)~\cite{sdxl_turbo}. 
Specifically, we train two discriminators. 
The first, $D_e$, receives an input frame, instruction, and output frame and attempts to determine if the edit was performed by the image editing teacher or video editing student. 
The second, $D_v$, receives a video and caption, and attempts to determine if the video was generated by the video generation teacher or video editing student.
We further follow ADD and employ the hinge loss~\cite{lim2017geometric} objective for adversarial training.
Hence, the discriminators minimize the following objectives:
\begin{gather*}
    \mathcal{L}_\text{D-Edit} = 
    \mathbb{E}_{x'_\psi}[\text{max}(0, 1-D_{e}(x'_\psi))] +\mathbb{E}_{x'_0}[\text{max}(0, 1+D_{e}(x'_0))], \\
    \mathcal{L}_\text{D-Video} = 
    \mathbb{E}_{x'_\rho}[\text{max}(0, 1-D_{v}(x'_\rho))] +\mathbb{E}_{x'_0}[\text{max}(0, 1+D_{v}(x'_0))], \\
    \label{eq:adv_dis}
\end{gather*}
while the student minimizes the following objectives:
\begin{gather*}
    \mathcal{L}_\text{G-Edit} = -\mathbb{E}_{x_0'}[\text{max}(0, 1+D_{e}(x_0'))], \\
    \mathcal{L}_\text{G-Video} = -\mathbb{E}_{x_0'}[\text{max}(0, 1+D_{v}(x_0'))],
    \label{eq:adv_gen}
\end{gather*}
where $x'_\psi$ and $x'_\phi$ are samples generated from random noise by applying the image editing and video generation teachers accordingly for multiple forward diffusion steps using DDIM sampling.

The combined loss to train our student model is:
\begin{gather*}
\mathcal{L}_\text{G-\method} = \alpha(\mathcal{L}_\text{G-Edit}+\lambda\mathcal{L}_\text{SDS-Edit}) + \beta(\mathcal{L}_\text{G-Video}+\lambda\mathcal{L}_\text{SDS-Video}), 
\label{eq:final_gen}
\end{gather*}
and the discriminators are trained with:
\begin{gather*}
\mathcal{L}_\text{D-\method} = \alpha\mathcal{L}_\text{D-Edit} + \beta\mathcal{L}_\text{D-Video}.
\label{eq:final_disc}
\end{gather*}
In practice, we set both $\alpha$ and $\beta$ to 0.5. Following~\cite{sdxl_turbo}, we set $\lambda$ to 2.5.

\subsection{Implementation Details}
\label{impl_details}
We provide below further implementation details on the timestep sampling during \method, and the architecture of our discriminators.
\vspace{-0.1cm}
\subsubsection{K-Bin Diffusion Sampling}
As previously mentioned, our student generates an edited video using $k$ diffusion steps, and we backpropagate the loss through all of the steps. 
During training we set $k=3$, as it is the maximum number of diffusion steps that fits into memory.
Notably, naively using the same $k$ timesteps during training, and setting a larger $k$ during inference time may lead to a train-test discrepancy.
To avoid such a train-test discrepancy we divide the $T$ diffusion steps into $k$ evenly sized bins, each containing $T/k$ steps. 
Then, during each training generation iteration, we randomly select a step from its corresponding bin.
We ablate this strategy in Sec.~\ref{sec:ablations}.
\vspace{-0.2cm}
\subsubsection{Discriminator Architecture}
The base architecture of our discriminators is similar to~\cite{sdxl_turbo}. Specifically, we utilize DINO~\cite{oquab2023dinov2} as a frozen feature network and add trainable heads to it. To add conditioning to the input image for $D_e$, we use an image projection in addition to the text and noisy image projection, and combine the conditions with an additional attention layer. To support video conditioning for $D_v$, we add a single temporal attention layer over the projected features of DINO, applied per pixel.
\vspace{-0.2cm}

\section{Experiments}
\label{sec:experiments}
\vspace{-0.2cm}
We perform a series of experiments to evaluate and analyze our approach.
We start by evaluating our method on the task of instruction-guided video editing. Specifically, we benchmark our video editing model, \fullmodel, versus multiple baselines on the Text-Guided Video Editing~(TGVE)~\cite{wu2023cvpr} benchmark.
Additionally, we expand TGVE with additional editing tasks, and benchmark our model against the expanded benchmark as well.
Then, we carry out an ablation study to measure the impact of different design choices that we introduced into our approach.
We continue by examining the capability of \model to perform zero shot video editing over tasks which were not presented during the alignment phase, but are part of the editing adapter's knowledge.
Finally, we conduct a qualitative study to verify that our approach can be applied to align other combinations of adapters.
Video editing examples are presented in Fig.~\ref{fig:method_sample}, with additional samples and qualitative comparisons available on the
supplementary material and website.\footnote{\url{https://fdd-video-edit.github.io/}}
\vspace{-0.1cm}
\subsubsection{Metrics.} The goal of text-based video editing is to modify a video in accordance with the input text, while preserving elements or aspects of the video that should not change. 
To this end, our evaluation in based on subjective and objective success metrics. 
We utilize two categories of objective metrics for evaluation. 
The first category includes metrics used by the TGVE competition: (i) CLIPFrame~(Frame Consistency)~\cite{wu2023cvpr} -- measuring the average cosine similarity among CLIP image embeddings across all video frames, and (ii) PickScore~\cite{pickapic} -- measuring the predicted human preference averaged over all video frames.

One inherent limitation of both metrics is their lack of consideration for temporal consistency. 
For example, CLIPFrame applies a simple average over a similarity score between images. 
As a result, CLIPFrame favours static videos with limited or no motion. 
When evaluating video-editing methods, output videos with little or no modification will score higher on this metric.

Therefore, we introduce additional metrics that makes use of ViCLIP~\cite{wang2023internvid}, a video CLIP~\cite{radford2021learning} model that considers temporal information when processing videos. 
Concretely, we add the following metrics: (i) ViCLIP text-video direction similarity~($\text{ViCLIP}_{dir}$, inspired by $\text{CLIP}_{dir}$~\cite{gal2021stylegan}) -- measuring agreement between change in captions and the change in videos, 
and (ii) ViCLIP output similarity~($\text{ViCLIP}_{out}$~\cite{radford2021learning}) -- measuring edited image similarity with output caption.

For subjective evaluation, we follow the TGVE~\cite{wu2023cvpr} benchmark and rely on human raters. 
We present the human raters with the input video, a caption describing the output video, and two edited videos. 
We then task the raters to answer the following questions: (i) Text alignment: Which video better matches the caption, (ii) Structure: Which video better preserves the structure of the input video, and (iii) Quality: Aesthetically, which video is better. 
In addition, we report an overall human evaluation score by averaging the preference score of all three questions.
\vspace{-0.2cm}
\subsubsection{Unsupervised Dataset}
Our \method approach requires a dataset with inputs for the student and teachers.
In the case of video editing, each data point contains $y = (c_{out}, c_{instruct}, c_{vid})$, where $c_{out}$ is an output video caption, $c_{instruct}$ is the editing instruction, and $c_{vid}$ is the input video. 
To create this dataset we utilize the videos from Emu Video's high-quality dataset, which has 1600 videos. 
For each video, we follow~\cite{emu-edit} and task Llama-2~\cite{touvron2023llama} with generating seven editing instructions, one for each of the following tasks from Emu Edit: Add, Remove, Background, Texture, Local, Style, Global.
\vspace{-0.2cm}
\subsubsection{Editing Tasks Unseen During Alignment}
\begin{figure}[t!]
\centering
\includegraphics[width=.80\linewidth]{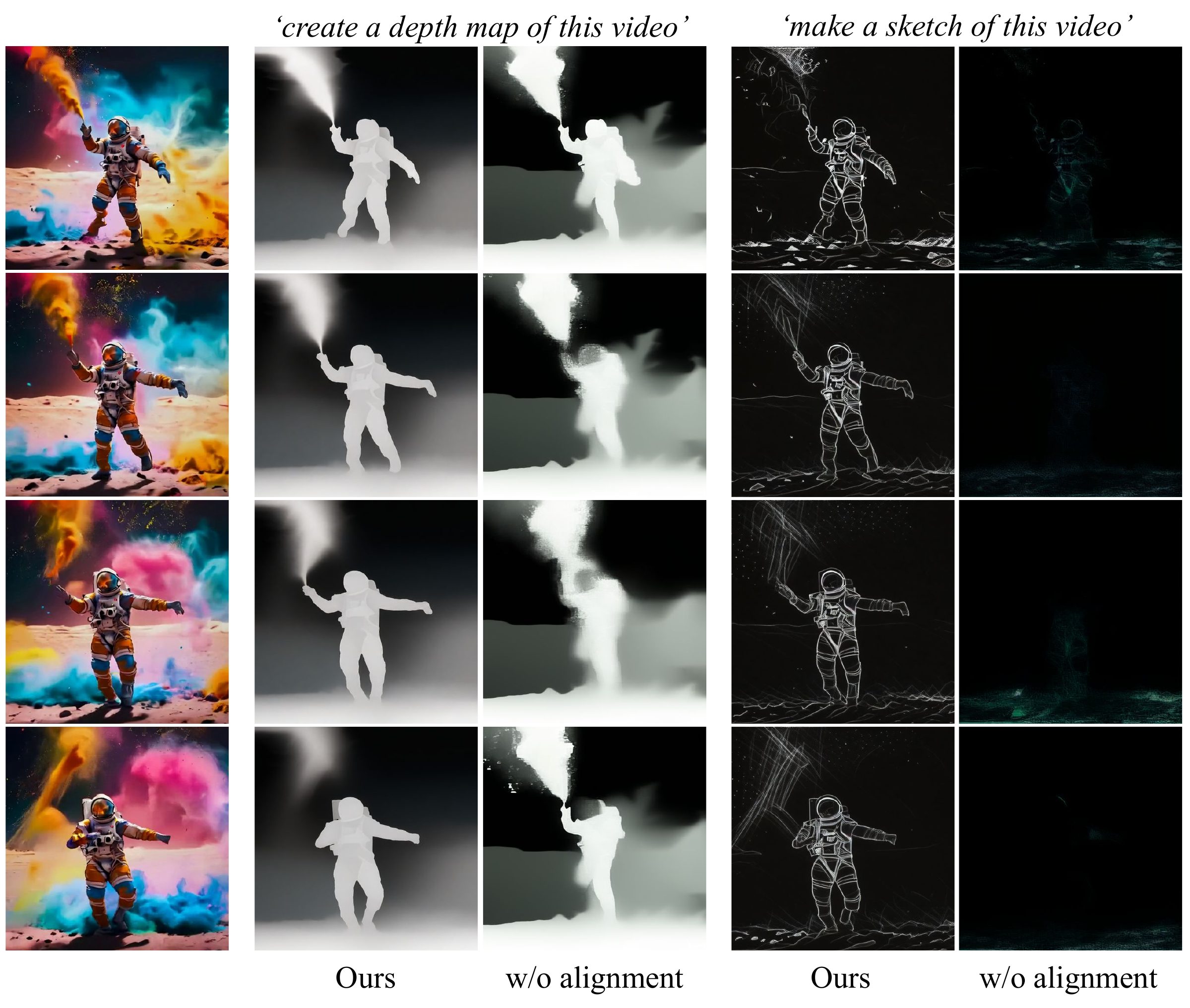}
\vspace{-0.1cm}
\caption{Our model performs zero-shot video editing for tasks that Emu Edit can execute on images, without explicitly training on them during alignment.}%
\label{fig:controlnet}
\vspace{-0.3cm}
\end{figure}

Throughout the alignment phase, we train our student model on a subset of the tasks that our image editing adapter was originally trained on.
For example, we do not train the student to segment objects in the input video, extract the pose, convert the video into a sketch, or derive depth maps from the input video.
However, we observe a significant improvement in student performance on these tasks following the alignment stage. %
This suggests that our student model aligns with the entire knowledge of the teacher model, even if we only expose it to a subset of this knowledge. 
We provide a qualitative comparison in Fig.~\ref{fig:controlnet} to illustrate the contribution of the alignment phase to the tasks mentioned above.
\vspace{-0.2cm}
\subsubsection{Training Details}
We train both adapters using the same frozen Emu backbone, with zero-terminal
SNR~\cite{zero_snr} enforced during training. We train the model for a total of 1500 iterations with a batch size of 64 and a fixed learning rate of 1e-5, without warm-up. For the first 1000 iterations we train only with the SDS losses, and in the subsequent 500 iterations we add the adversarial losses. We train on 8 frame video clips with a resolution of 512 $\times$ 512. Throughout the paper we generate examples using the Denoising Diffusion Implicit Models~(DDIM) algorithm. We condition the edit adapter on the task label following~\cite{emu-edit}, and the video adapter on the first frame following~\cite{emu-video}. Specifically, we edit the first frame using the edit adapter. To generate videos longer than 8 frames, we follow the same procedure as~\cite{instructvid2vid} and apply a sliding window over the input video.

\subsection{Video Editing Benchmark}
\label{sec:benchmark}

At present, the TGVE~\cite{wu2023cvpr} benchmark is the established standard for evaluating text-based video editing methods.
The benchmark contains seventy six videos, and each has four editing prompts. 
All videos are either 32 or 128 frames with a resolution of 480$\times$480. 
The benchmark encompasses four types of editing tasks: (i) local object modification, (ii) style change, (iii) background change, and (iv) execution of multiple editing tasks simultaneously.

Due to TGVE's focus on a narrow range of editing tasks we choose to increase its diversity, by adding three new editing tasks: (i) object removal (Remove), (ii) object addition (Add), and (iii) texture alterations (Texture). 
For each video in TGVE and for each new editing operation, we assign crowd workers the task of writing down an editing instruction and an output caption describing a desired output video.
In the remainder of this section, we report the performance of our method on the TGVE benchmark and our extension thereof, which we name TGVE+.
In the supplementary material, we provide examples of our benchmark, which we are publicly releasing to support ongoing research in video editing.
\vspace{-0.1cm}
\subsection{Baseline Comparisons}
\label{sec:baselines}
\vspace{-0.1cm}
\begin{table*}[t]
\centering
\caption{Comparison with video-editing baselines on the TGVE and TGVE+ benchmarks. We report PickScore, CLIPFrame, ViCLIP metrics and human ratings. Human evaluation shows the percentage of raters that prefer the results of \model.}
\vspace{-0.1cm}
\label{tab:eval_video_editing_tgve}
\scalebox{0.83}{
\centering
\begin{tabular}{l|lcccc|cccc}
\toprule
Dataset & Method & $\text{PickScore}\!\uparrow$ & $\text{CLIPFrame}\!\uparrow$ &  $\text{ViCLIP}_{dir}\uparrow$ & $\text{ViCLIP}_{out}\!\uparrow$ & Text & Struct. &  Quality & Avg. \\
\midrule
\multirow{ 4}{*}{TGVE} &
TAV~\cite{wu2023tune}    & 20.36 & 0.924 & 0.162 & 0.243 & 72.4 & 74.0 & 85.2 & 77.2 \\
& SDEdit~\cite{sdedit} & 20.18 & 0.896 & 0.172 & 0.253 & 75.7 & 67.4 & 79.0 & 74.0 \\
& STDF~\cite{yatim2023space}   & 20.40 & \textbf{0.933} & 0.110 & 0.226 & 81.3 & 65.8 & 70.1 & 72.4 \\
& Fairy~\cite{Wu2023FairyFP}   & 19.80 & \textbf{0.933} & 0.164 & 0.208 & 77.3 & 62.8 & 75.0 & 71.7 \\

& InsV2V~\cite{instructvid2vid} & \textbf{20.76} & 0.911 & 0.208 & \textbf{0.262} & 57.9 & 55.9 & 65.1 & 59.6 \\
& \model (Ours) & \textbf{20.76} & 0.922 & \textbf{0.221} & \textbf{0.262} & -- & -- & -- & -- \\
\midrule
\multirow{4}{*}{\shortstack[c]{TGVE+}} & TAV~\cite{wu2023tune}    & 20.47 & \textbf{0.933} & 0.131 & 0.242 & 72.2 & 74.0 & 77.2 & 74.5 \\
& SDEdit~\cite{sdedit} & 20.35 & 0.899 & 0.144 & 0.246 & 74.5 & 68.5 & 77.9 & 73.6 \\
& STDF~\cite{yatim2023space}   & 20.60 & \textbf{0.933} & 0.093 & 0.227 & 78.6 & 70.2 & 72.6 & 73.8 \\
& Fairy~\cite{Wu2023FairyFP}   & 19.81 & \textbf{0.933} & 0.140 & 0.197 & 74.4 & 70.8 & 77.8 & 74.3 \\
& InsV2V~\cite{instructvid2vid} & 20.37 & 0.925 & 0.174 & 0.236 & 62.9 & 56.4 & 61.4 & 60.2 \\
& \model (Ours) & \textbf{20.88} & 0.926 & \textbf{0.198} & \textbf{0.251} & -- & -- & -- & -- \\
\bottomrule
\end{tabular}}
\vspace{-0.1cm}
\end{table*}
\begin{table*}[t]
\centering
\caption{Ablation study on our different contributions. We report human ratings on video-editing results on the TGVE benchmark. Human evaluation shows the percentage of raters that prefer the results of \model.}
\label{tab:abl_study}
\vspace{-0.1cm}
\scalebox{1.0}{
\centering
\begin{tabular}{l@{\hspace{1cm}}cccc}
\toprule
Method & Text & Struct. &  Quality & Avg. \\
\midrule
Random Init              & 96.7 & 70.1 & 94.7 & 87.2 \\
w/o alignment            & 77.6 & 91.4 & 89.8 & 86.3 \\
w/o SDS                  & 77.6 & 87.5 & 92.1 & 85.7 \\
w/o discriminators       & 74.3 & 84.2 & 83.9 & 80.8 \\
w/o K-Bin Sampling       & 57.6 & 49.7 & 51.6 & 53.0 \\
\bottomrule
\end{tabular}}
\vspace{-0.3cm}
\end{table*}
We benchmark our model against InsV2V~\cite{instructvid2vid}, which is the leading performer in the TGVE benchmark. For completeness, we also compare to Space-Time Diffusion Features (STDF)~\cite{yatim2023space}, which is one of the latest motion transfer works, Tune-A-Video (TAV)~\cite{wu2023tune}, which served as the baseline in the TGVE contest, SDEdit~\cite{sdedit} which is a popular diffusion editing baseline, and Fairy~\cite{Wu2023FairyFP}. For SDEdit, we use a noising level of $0.75$ after comparing multiple noising levels and choosing the best one with respect to automatic metrics.
Unlike the official TGVE contest, which compared all participating methods to TAV, we directly compare our model with the different baselines.
 
\begin{figure}[t!]
\centering
\includegraphics[width=0.9\linewidth]{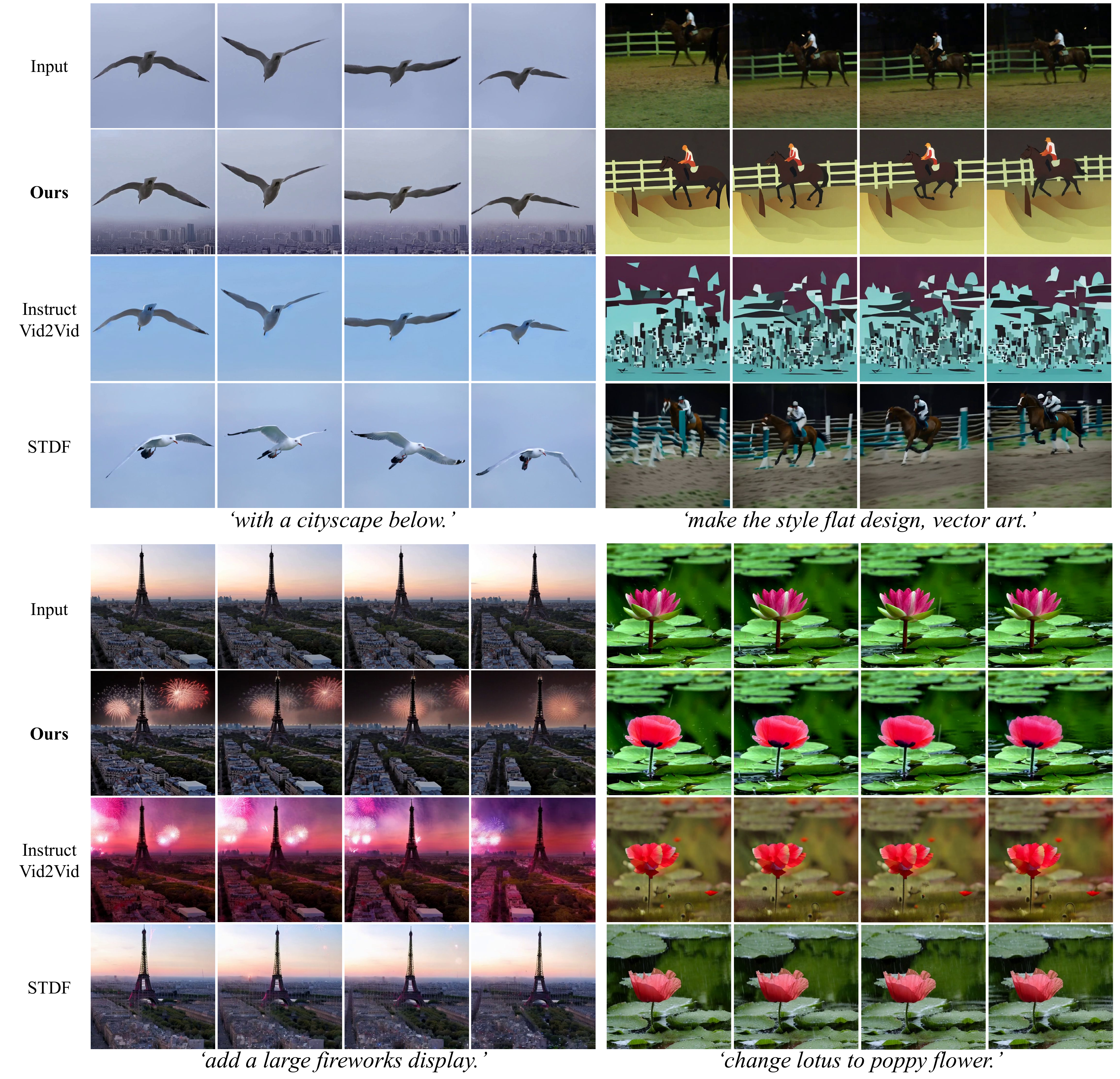}
\vspace{-0.1cm}
\caption{Comparison of our model against baselines using examples from the Text-Guided Video Editing~(TGVE)~\cite{wu2023cvpr} benchmark and our extension of it.} 
\label{fig:qualitative_tgve}
\vspace{-0.5cm}
\end{figure}

Tab.~\ref{tab:eval_video_editing_tgve} shows our results versus the baselines.
As can be seen, human raters prefer \model over all baselines by a significant margin.
Moreover, when considering automatic metrics, \model presents state-of-the-art results on all of the objective metrics except for CLIPFrame.
Interestingly, while STDF and Fairy achieve the highest scores on the CLIPFrame metric, human raters prefer our model in 72.4\% and 71.7\% of the time respectively.
In addition to numerical results, Fig.~\ref{fig:qualitative_tgve} provides a visual comparison between the outputs of \model and top performing baselines.

\begin{figure}[t!]
\centering
\includegraphics[width=0.98\linewidth]{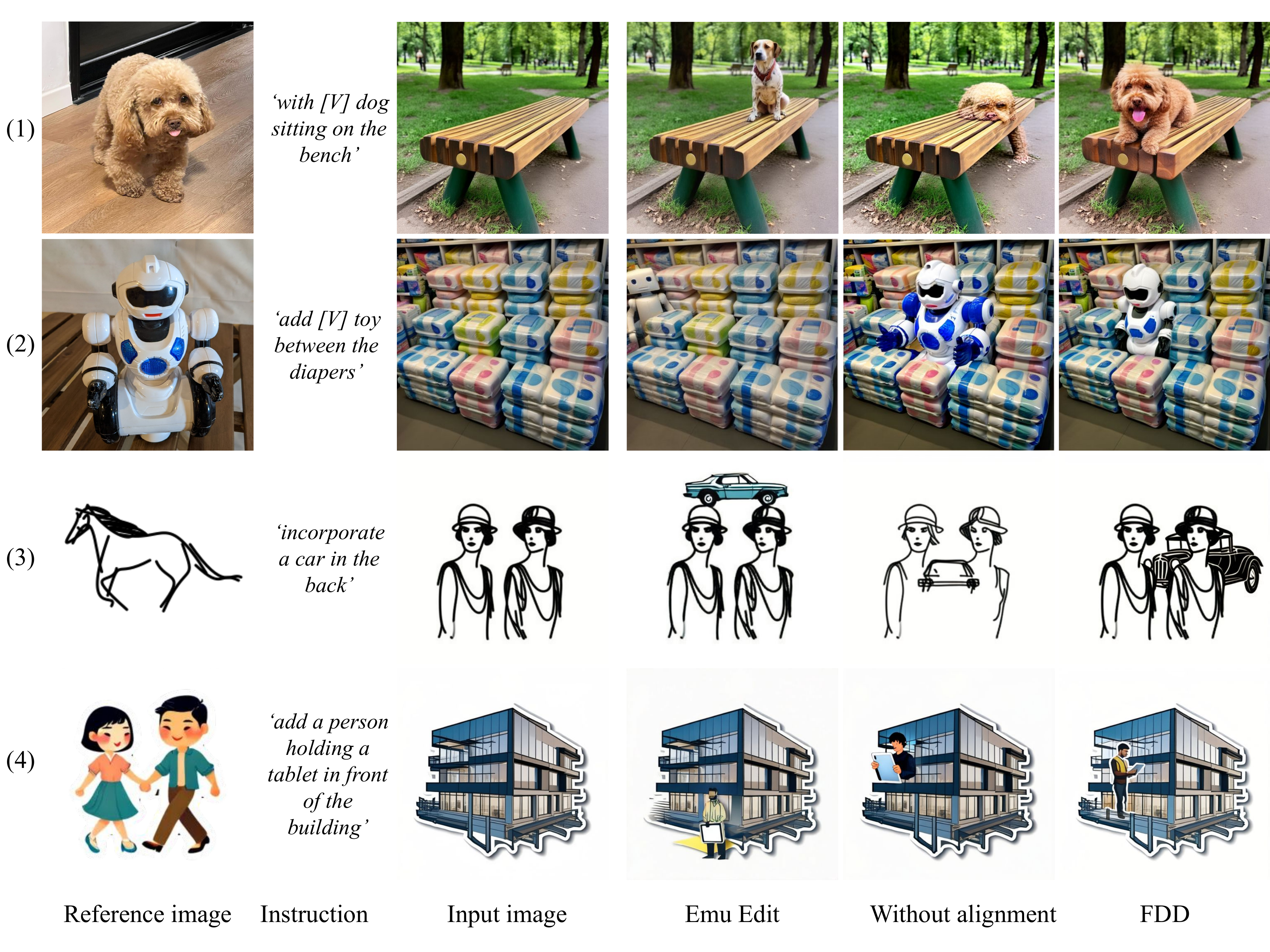}
\caption{We apply \method to combine an editing adapter with LoRA-based adapters: (1) a subject-driven adapter for a dog, (2) a subject-driven adapter for a toy robot, (3) a style-driven adapter for line art, and (4) a style-driven adapter for stickers.}
\vspace{-0.2cm}
\label{fig:additional_combination}
\vspace{-0.4cm}
\end{figure}

\subsection{Ablations}
\label{sec:ablations}
We provide an ablation study of human ratings in Tab.~\ref{tab:abl_study} to assess the effectiveness of our different contributions on the TGVE+ benchmark. 
We begin our study by ablating our decision to add the pre-trained adapters to our student model rather than learning them jointly during the alignment procedure. 
In this experiment~(Random Init), we initialize the ControlNet edit adapter with the weights from the text-to-image encoder, and the temporal layers are initialized to identity. We then fine-tune the entire resulting model.
Our observation indicates that this particular variant is unsuccessful in acquiring proficiency in the video-editing task, implying that \method is more adept at aligning pre-trained adapters rather than initiating their training from scratch.

We continue by ablating the design of the alignment procedure itself, examining three methods of combining the adapters: (i) without any alignment~(w/o alignment), (ii) using only the adversarial loss and excluding SDS~(w/o SDS), and (iii) incorporating SDS but excluding the adversarial loss~(w/o Discriminators). 
As expected, using no alignment leads to poor results for both structure preservation and quality aspects. 
This suggests that \method is essential when combining adapters that were trained separately for different tasks.
When evaluating the contribution of each term in \model, namely SDS and adversarial loss, the SDS term has a larger impact on the alignment process.
Interestingly, employing the adversarial term alone is sufficient to achieve some level of alignment. 
However, the use of both terms is vital for successful alignment. \\
We conclude by validating the contribution of using K-Bin diffusion sampling.
For this ablation, we sample $k$ steps uniformly throughout training instead of randomly sampling them from $k$ buckets. 
As evident from the results, the process of sampling steps from $k$ bins further enhances the performance of \method.

\vspace{-0.2cm}
\subsection{Additional Combinations of Adapters}
\label{sec:generalization}
In this section we explore the ability of \method to align other adapters.
To this end, we train four different LoRA adapters on our text-to-image backbone; two for subject-driven generation~\cite{ruiz2023dreambooth} and two for style-driven generation. 
We then align each of them with our image editing adapter to facilitate personalized and stylized image editing capabilities. 

To create an unsupervised dataset for stylized editing, we utilize 1,000 (input caption, instruction, output caption) triplets from Emu Edit's dataset.
For personalized editing, we use 1,000 input captions and task Llama-2 with generating instructions for adding the subject or replacing items in the image with the subject.
Notably, we do not use images during training and instead generate the input images using the LoRA adapter. While each LoRA adapter requires a different alignment, we note that one can use a subject-conditioned adapter such as ReferenceNet~\cite{hu2023animate} and perform a single alignment for all subjects and styles at once.

In Fig~\ref{fig:additional_combination}, we present qualitative examples of our method when applied to these combinations. For each input image and instruction we display the samples obtained by using: (i) vanilla Emu Edit, (ii) attaching both adapters without alignment, and (iii) after alignment. 
As anticipated, Emu Edit is not capable of personalized editing as it lacks awareness of the desired subject. Similarly, for stylized editing, it has difficulty maintaining the style of the input. When using a ``plug-and-play'' approach, the model either fails to maintain the style or subject identity, or it produces unsatisfactory generations with significant artifacts. However, after alignment, the edits become more consistent with the reference style and subject.
\vspace{-0.2cm}

\section{Limitations}
\label{sec:limitations}
\vspace{-0.2cm}
We identify two main limitations in our approach. 
Fundamentally, the performance of our model is upper bounded by the capabilities of the different teacher models. 
For instance, we anticipate that the editing capabilities of our student model will not exceed those of the image editing adapter teacher.
Secondly, our method is intended to align a student which is initialized with pre-trained adapters. As we show in Sec.~\ref{sec:ablations}, training it from scratch results in poor performance. Thus, it requires the teachers to be trained as adapters over a frozen text-to-image backbone. 
We hope that future work will overcome these limitations to enable more efficient training and better performing models.
\vspace{-0.2cm}

\section{Conclusions}
\label{sec:conclusions}
\vspace{-0.2cm}
In this study, we proposed an approach for learning how to edit videos without supervised video editing data.
Our approach is based on the key insight that video editing can be factorized into two distinct criteria: (1) precise editing of video frames, and (2) ensuring temporal consistency among the edited frames. 
Leveraging this insight, we proposed a two-stage approach for training a video editing model.
In the first stage, we separately train an image editing adapter and a video generation adapter, and attach both to a frozen text-to-image backbone.
Subsequently, we align the adapters towards video editing using \fullmethod (\method). 
Our results demonstrate that the resulting model, \fullmodel, establishes a new state-of-the-art in video editing.
Finally, we show the potential of our approach for learning other tasks by aligning additional adapters. 
There are still many opportunities for applying our approach to other tasks, and alleviating some of its limitations, and we are excited to see how the research community will utilize and build upon this work in the future.

\ifreview
\else
\section*{Acknowledgements}   
Andrew Brown,
Bichen Wu,
Ishan Misra,
Saketh Rambhatla,
Xiaoliang Dai,
Zijian He. Thank you for your contributions!
\fi

\bibliographystyle{splncs04}
\bibliography{main}
\end{document}